%% file: emnlp2017.tex
\documentclass[11pt,letterpaper]{article}
\usepackage{emnlp2017}
\usepackage{times}
\usepackage{latexsym}
\usepackage{algorithm2e}
\usepackage{multirow}
\usepackage{graphicx}
\usepackage{amsmath}
\usepackage{amssymb}
\usepackage{microtype}
\usepackage[hang,flushmargin]{footmisc}   
\usepackage{url}

\emnlpfinalcopy

\def\emnlppaperid{8}

\title{Cross-genre Document Retrieval:\\ Matching between Conversational and Formal Writings}

\author{Tomasz Jurczyk \\ Mathematics and Computer Science \\  Emory University \\ Atlanta, GA 30322, USA \\ \tt{tomasz.jurczyk@emory.edu}
         \And
         Jinho D. Choi \\ Mathematics and Computer Science \\  Emory University \\ Atlanta, GA 30322, USA \\ \tt{jinho.choi@emory.edu}}

\date{}

\begin{document}

\maketitle

\input{abstract.tex}
\input{introduction.tex}
\input{related.tex}
\input{data.tex}
\input{approach.tex}
\input{experiments.tex}
\input{conclusion.tex}

\bibliography{emnlp2017}
\bibliographystyle{emnlp_natbib}

\end{document}

%% file: abstract.tex
\begin{abstract}
This paper challenges a cross-genre document retrieval task, where the queries are in formal writing and the target documents are in conversational writing.
In this task, a query, is a sentence extracted from either a summary or a plot of an episode in a TV show, and the target document consists of transcripts from the corresponding episode.
To establish a strong baseline, we employ the current state-of-the-art search engine to perform document retrieval on the dataset collected for this work.
We then introduce a structure reranking approach to improve the initial ranking by utilizing syntactic and semantic structures generated by NLP tools. 
Our evaluation shows an improvement of more than 4\% when the structure reranking is applied, which is very promising.

\end{abstract}

%% file: introduction.tex
\section{Introduction}

Document retrieval has been a central task in natural language processing and information retrieval.
The goal is to match a query against a set of documents.
Over the last decade, advanced techniques have emerged and provided powerful systems that can accurately retrieve relevant documents~\cite{blair1985evaluation,callan1994passage,cao2006adapting}.
While the retrieval part is crucial, proper ranking of the retrieved documents can significantly improve the overall user satisfation by putting more relevant documents at the top~\cite{balinski2005re,yang2006document,zhou2009latent}.
Many previous works provide strong baselines for unstructured text retrieval and ranking problems; however, these systems usually assume a homogeneous domain for queries and target documents.

Due to the spike of applications that are required to maintain the conversation, dialog data has recently become a popular target among researchers. The work in this field concerns problems such as learning facts through conversation~\cite{fernandez2011reciprocal,williams2015rapidly,hixon2015learning} or dialog summarization~\cite{oya-carenini:2014:W14-43,misra-EtAl:2015:NAACL-HLT}. More recent work in this field has focused on several inter-dialogue tasks~\cite{xu-reitter:2016:P16-1,kim-banchs-li:2016:P16-1,he-EtAl:2016:P16-12}.
To the best of our knowledge our work is the first, where the cross-genre document retrieval is analyzed based on conversational and formal writings.

This paper analyzes the performance of state-of-the-art retrieval techniques targeting TV show transcripts and their descriptions.
We first collect a dataset comprising transcripts from a popular TV show and their summaries and plots (Section~\ref{data-section}).
We then establish a solid baseline by adapting an advanced search engine and implement structure reranking to improve the initial ranking from the search engine (Section~\ref{approach-section}).
Our evaluation shows a 4\% improvement, which is significant (Section~\ref{experiments-section}).

%% file: related.tex
\section{Related work}
Information extraction for dialogue data has already been widely explored.
~\newcite{yoshino-mori-kawahara:2011:SIGDIAL2011} presented a spoken dialogue system that extracts predicate-argument structures and uses them to extract facts from news documents.
~\newcite{FlychtEriksson-Joensson:2003:SIGDIAL} developed a dialogue interaction process of accessing textual data from a bird encyclopedia. 
An unsupervised technique for meeting summarization using decision-related utterances has been presented by~\newcite{wang-cardie:2012:SIGDIAL20122}.
~\newcite{gorinski-lapata:2015:NAACL-HLT} studied movie script summarization. 
All the aforementioned work uses the syntactic and semantic relation extraction and thus is similar to ours; however, it is distinguished in a way that it lacks a cross-genre aspect. 

\begin{table*}[htbp]
\centering\small
\begin{tabular}{c|l||l}
\multicolumn{2}{c||}{\textbf{Dialogue}} & \multicolumn{1}{c}{\textbf{Summary + Plot}} \\ \hline\hline
Joey & \begin{tabular}[c]{@{}l@{}}One woman? That's like saying there's only one flavor \\ of ice cream for you. Lemme tell you something, Ross. \\ There's lots of flavors out there.\end{tabular} & Joey compares women to ice cream.~(\textit{S}) \\ \hline
Ross & \begin{tabular}[c]{@{}l@{}}You know you probably didn't know this, but back in \\ high school, I had a, um, major crush on you.\end{tabular} & \multirow{2}{*}{\begin{tabular}[c]{@{}l@{}}Ross reveals his high school crush on Rachel.~(\textit{S}) \end{tabular}} \\
Rachel & I knew. &  \\ \hline
Chandler & Alright, one of you give me your underpants. & Chandler asks Joey for his underwear, but \\
Joey     & Can't help you, I'm not wearing any.         & Joey can't help him out as he's not wearing any.~(\textit{P})
\end{tabular}
\caption{\small{Three manually curated examples of dialogues and their descriptions.}}
\label{tbl:examples}
\vspace{-2ex}
\end{table*}

%% file: data.tex
\section{Data}
\label{data-section}

The Character Mining project provides transcripts of the TV show, \textit{Friends}; transcripts from 8 seasons of the show are publicly available in the JSON format,\footnote{\url{nlp.mathcs.emory.edu/character-mining}} where the first 2 seasons are annotated for the character identification task~\cite{chen-choi:2016:SIGDIAL}.
Each season consists of episodes, each episode contains scenes, each scene includes utterances, where each utterance comes with the speaker information.

For each episode, the episode summary and plot are first collected from fan sites,\footnote{\url{friends-tv.org}, \url{friends.wikia.com}} then sentence segmented by NLP4J,\footnote{\url{github.com/emorynlp/nlp4j}} the same tool used for the provided transcripts.
Generally, summaries give broad descriptions of the episodes, whereas plots describe facts within individual scenes.
Finally, we create a dataset by treating each sentence as a query and its relevant episode as the target document.
Table~\ref{table:query-data-stat} shows the distributions of this dataset.

\begin{table}[htbp]
\centering\small
\begin{tabular}{l|r||l|r}
\multicolumn{2}{c||}{\textbf{Dialogue}} & \multicolumn{2}{c}{\textbf{Summary + Plot}} \\ \hline\hline
\# of episodes &     194 & \# of queries &   5,075 \\
\# of tokens   & 897,446 & \# of tokens  & 119,624 \\
\end{tabular}
\caption{Dialogue, summary, and plot data.}
\label{table:query-data-stat}
\vspace{-2ex}
\end{table}



%% file: approach.tex
\section{Structure Reranking}
\label{approach-section}

For each query (summary or plot) in the dataset, the task is to retrieve the document (episode) most relevant to the query.
The challenge comes from the cross-genre aspect: how to retrieve documents in dialogues given the queries in formal writing.
This section describes our structure reranking approach that significantly outperforms an advanced search engine, Elasticsearch\footnote{\url{https://www.elastic.co/}}.


\subsection{Relation Extraction}
\label{subsection-structure-extraction}

Since our queries and documents appear very different on the surface level (Table~\ref{tbl:examples}), relations are first extracted from them and matching is performed on the relation level, which abstracts certain pragmatic differences between these two types of writings.
All data are lemmatized, tagged with parts-of-speech and named entities, parsed into dependency trees, and labeled with semantic roles using NLP4J.

A sentence may consist of multiple predicates, and each predicate comes with a set of arguments.
A predicate together with its arguments is considered a relation.
For each argument, heuristics are applied to extract meaningful contextual words by traversing the subtree of the argument.
Our heuristics are designed for the type of dependency trees generated by NLP4J, but similar rules can be generalized to other types of dependency trees.
Relations from dialogues are attached with the speaker names to compensate the lack of entity information.

\begin{figure}[htbp!]
\centering
\includegraphics[scale=0.29]{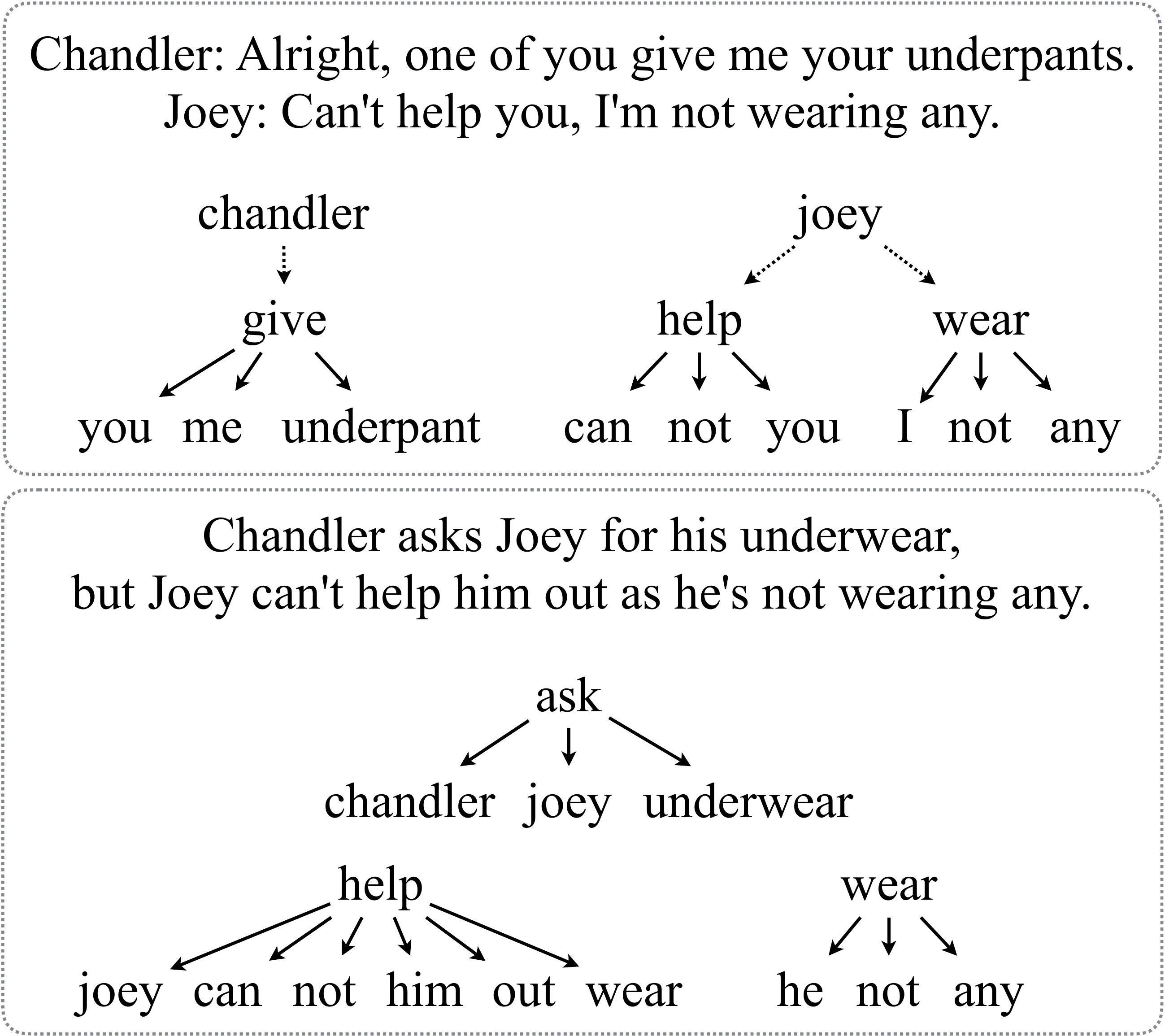}
\caption{\small{Two sets of relations, from dialogue and plot, extracted from the examples in Table~\ref{tbl:examples}}.}
\label{fig:relation-extraction}
\end{figure}

\begin{figure*}[htbp!]
\centering
\includegraphics[scale=0.45]{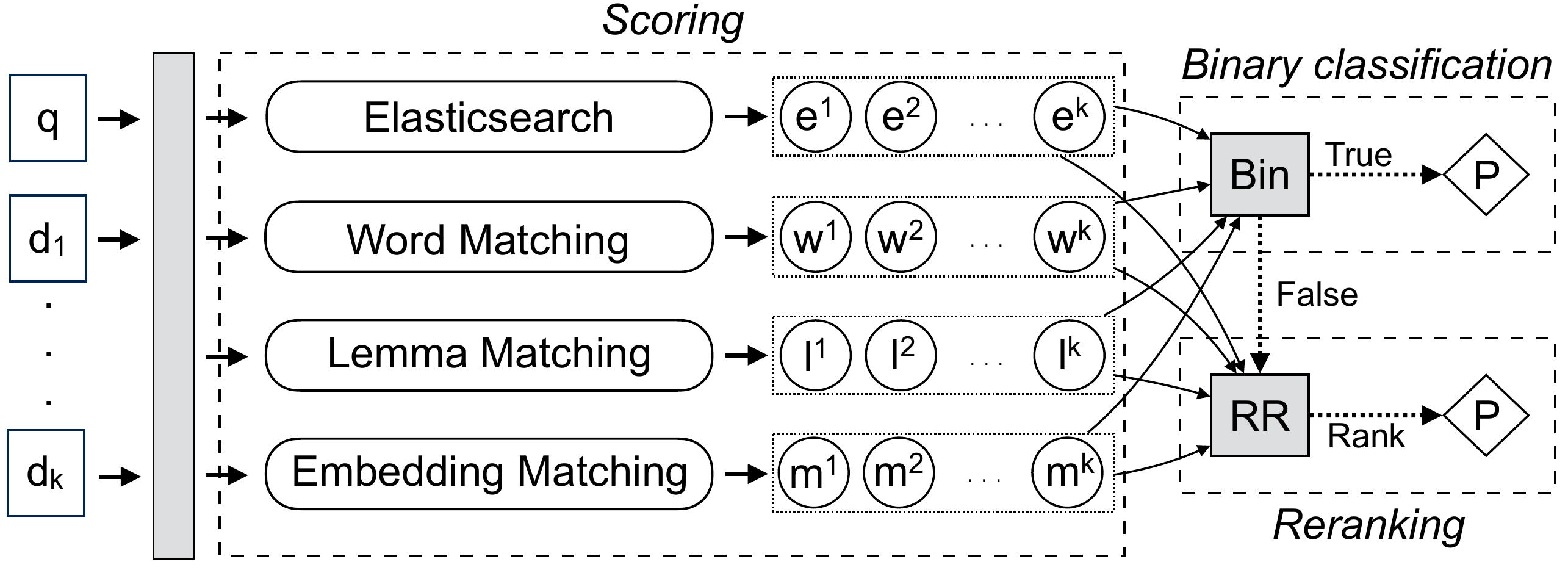}
\caption{The overview of our structure reranking. Given documents $d_1,\ldots,d_k$ and a query $q$, 4 sets of scores are generated: the Elasticsearch scores and the matching scores using 3 comparators: \textit{word}, \textit{lemma}, and \textit{embedding}. The binary classifier \texttt{Bin} predicts whether the highest ranked document from Elasticsearch is the correct answer. If not, the system \texttt{RR} reranks the documents using all scores and returns a new top-ranked prediction.}
\label{figure:system-overview}
\vspace{-2ex}
\end{figure*}

\noindent By extracting relations that comprise only meaningful words, it prunes out much noise (e.g., disfluency), which allows the system to retrieve relevant documents with higher precision.
While our relation extraction is based on the sentence level, it can be extended to the document level by adding coreference relations, which we will explore in the future.

%
%
%
%
%
%

\subsection{Structure Matching}
\label{ssec:structure-matching}

All relations extracted from dialogues are stored in an inverted index manner, where words in each relation are associated with the relation and the episode in which the relation occurs.
Algorithm~\ref{algorithm:scoring-algorithm} shows how our structure matching works.
Given a list of documents retrieved from the index based on a query $q$, it first initializes scores for all documents to 0.
For each document $d_i$, it compares each relation $r^q$ from $q$ to relations extracted from $d_i$.
The relation $r$ from $d_i$ is kept within $R^d$ if it has at least one word that overlaps with $r^q$.
For each relation $r^d \in R^d$, the comparator function returns the matching score between $r^d$ and $r^q$.
The maximum matching score is added to the overall score of this document.
This procedure is repeated; finally, the algorithm returns the overall matching scores for all documents.

\begin{algorithm}[htbp!]
\small 
  \KwIn{$D$: a list of documents, $q$: a query.\linebreak $f_r$: a function returning all relations.\linebreak $f_c$: a comparator function.}
 \KwOut{$S$: a list of matching scores for $D$.}
   $S \leftarrow$ [$0$ for $i \in [1, |S|]$\\
 \ForEach{$d_i \in D$}{
    \ForEach{$r^q \in f_r(q)$}{
      $R^d \leftarrow$ [$r$ for $r \in f_r(d_i)$ if $|r \cap r^q| \geq 1$]\\
      $s_{m} \leftarrow 0$\\    
      \ForEach{$r^d$ in $R^d$}{
        $s \leftarrow$ $f_c(r^d, r^q)$\\
        $s_{m} \leftarrow \max(s_{m}, s)$}
      $S_i \leftarrow$ $S_i + s_m$
    }
  }
 \caption{The structure matching algorithm.}
 \label{algorithm:scoring-algorithm}
\end{algorithm}

\noindent The comparator function $f_c$ takes two relation sets, $r^d$ and $r^q$, and returns the matching score between those two sets.
For $word$ and $lemma$, the count of overlapping words between them is used to produce two scores, $r^d_s$, and $r^q_s$, normalized by the length of the utterance and the query, respectively.
The harmonic mean of the two scores is then returned as the final score.
For $embedding$, $f_c$ uses embeddings to generate sum vectors from both sets and returns the cosine similarity of these two vectors.

\subsection{Document Reranking}
\label{subsection-document-reranking}

The Elasticsearch scores and the 3 sets of matching scores for the top-$k$ documents (ranked by Elasticsearch) are fed into a binary classifier to determine whether or not to accept the highest ranked document.
A Feed Forward Neural Network with one hidden layer of size 15 is used for this classification.
If the binary classifier disqualifies the top-ranked document, the top-$k$ documents are reranked by the weighted sums of these scores.
A grid search is performed on the development set to find the optimized set of the weights.
At last, the system returns the document with the highest reranked scores:\\$d_i = \arg\max_i(\lambda_e\cdot e_i + \lambda_w\cdot w_i + \lambda_l\cdot l_i + \lambda_m\cdot m_i)$.

%% file: experiments.tex
\section{Experiments}
\label{experiments-section}

The data in Section~\ref{data-section} is split into training, development and evaluation sets, where queries from each episode are randomly assigned.
Two standard metrics are used for evaluation: precision at $k$ (P@k) and mean reciprocal rank (MRR).

\begin{table}[htbp!]
\centering\small
\begin{tabular}{l||r|r||r}
\multicolumn{1}{c||}{\bf Dataset} & \multicolumn{1}{c|}{\bf Summary} & \multicolumn{1}{c||}{\bf Plot} & \multicolumn{1}{c}{\bf Total} \\
\hline\hline
Training    & 970 & 3,013 & 3,983 (78.48\%) \\ 
Development &  97 &   403 &   500 $\:\:$(9.85\%) \\ 
Evaluation  & 150 &   442 &   592 (11.67\%) \\
\end{tabular}
\caption{Data split (\# of queries).}
\label{tbl:data-split}
\end{table}

\begin{table*}[htbp!]
\centering
\small
\begin{tabular}{l||c|c||c|c||c|c||c|c||c|c||c|c}
\multicolumn{1}{c||}{\multirow{3}{*}{\bf Model}} & \multicolumn{6}{c||}{\bf Development} & \multicolumn{6}{c}{\bf Evaluation} \\ \cline{2-13}
 & \multicolumn{2}{c||}{\bf Summary} & \multicolumn{2}{c||}{\bf Plot} & \multicolumn{2}{c||}{\bf All} & \multicolumn{2}{c||}{\bf Summary} & \multicolumn{2}{c||}{\bf Plot} & \multicolumn{2}{c}{\bf All} \\ \cline{2-13} 
                   & \bf P@1   & \bf MRR   & \bf P@1   & \bf MRR   & \bf P@1   & \bf MRR   & \bf P@1   & \bf MRR   & \bf P@1   & \bf MRR   & \bf P@1   & \bf MRR   \\ \hline \hline
Elastic$_{10}$     &     44.33 &     53.64 &     46.40 &     54.97 &     46.00 &     54.71 &     50.67 &     60.87 &     46.61 &     55.06 &     47.64 &     56.53 \\ \hline
Struct$_{w}$       &     38.14 &     48.42 &     34.00 &     45.11 &     34.80 &     45.75 &     35.33 &     48.34 &     35.52 &     47.08 &     35.47 &     47.40 \\
Struct$_{l}$       &     39.18 &     49.24 &     34.74 &     46.29 & \bf 35.60 & \bf 46.86 &     44.00 &     55.55 &     38.01 &     49.24 & \bf 39.53 & \bf 50.84 \\
Struct$_{m}$       &     35.05 &     46.71 &     33.50 &     44.72 &     33.80 &     45.10 &     36.00 &     50.14 &     35.97 &     46.95 &     35.98 &     47.76 \\ \hline
Rerank$_{1}$       &    47.42  &    55.66  &     48.39 &     56.10 &     48.20 &     56.02 &     56.67 &     63.77 &     50.23 &     57.99 &     51.86 &     59.46     \\
Rerank$_{\lambda}$ &     50.52 &     57.66 &     51.36 &     57.76 & \bf 51.20 & \bf 57.74 &     55.33 &     63.88 &     50.90 &     58.47 & \bf 52.03 & \bf 59.84
\end{tabular}
\caption{Evaluation on the development and evaluation sets for summary, plot, and all (summary + plot). Elastic$_{10}$: Elasticsearch with $k$ = 10, Struct$_{w,l,m}$: structure matching using words, lemmas, embeddings, Rerank$_{1,\lambda}$: unweighted and weighted reranking.}
\label{table:model-results}
\end{table*}

\subsection{Elasticsearch}

Elasticsearch is used to establish a strong baseline.\footnote{\url{www.elastic.co/products/elasticsearch}}
Each episode is indexed as a document using the default setting, Okapi BM25~\cite{robertson2009probabilistic}, and the TF-IDF based similarity with improved normalization; the top-$k$ most relevant documents are retrieved for each query.
While P@1 is less than 50\% (Table~\ref{table:elastic-precisionk-mrr}), P@10 shows greater than 70\% coverage implying that it is possible to achieve a higher P@1 by reranking results from $k \geq 10$.


\begin{table}[htbp!]
\centering\small
\begin{tabular}{r||c|c||c|c}
\multirow{2}{*}{\bf k} & \multicolumn{2}{c||}{\bf Development} & \multicolumn{2}{c}{\bf Evaluation} \\
\cline{2-5} 
   & \bf P@k   & \bf MRR   & \bf P@k   & \bf MRR     \\
\hline \hline
 1 & \bf 46.00 &     46.00 & \bf 47.64 &     47.64   \\
 5 &     65.80 &     53.80 &     69.26 &     69.26   \\
10 &     72.60 & \bf 54.71 &     74.66 & \bf 56.53   \\
20 &     78.80 &     55.13 &     79.73 &     56.91   \\
40 &     83.80 &     55.31 &     84.80 &     57.08   \\    
\end{tabular}
\caption{Elasticsearch results on (summary + plot).}
\label{table:elastic-precisionk-mrr}
\end{table}

%

\subsection{Structure Matching}

The Struct$_{*}$ rows in Table~\ref{table:model-results} show the results based on structure matching (Section~\ref{ssec:structure-matching}).
The highest P@1 of 39.53\% is achieved on the evaluation set using lemmas.
Although it is about 8\% lower than the one achieved by Elasticsearch, we hypothesize that this approach can correctly retrieve documents for certain queries that Elasticsearch cannot.

\begin{table}[htbp!]
\centering\small
\begin{tabular}{l||c|c||c|c}
\multicolumn{1}{c||}{\multirow{2}{*}{\bf Model}} & \multicolumn{2}{c||}{\bf Development} & \multicolumn{2}{c}{\bf Evaluation} \\ \cline{2-5} 
 & \multicolumn{1}{c|}{\bf P@1} & \multicolumn{1}{c||}{\bf MRR} & \multicolumn{1}{c|}{\bf P@1} & \multicolumn{1}{c}{\bf MRR} \\ \hline \hline
Elastic$_{10}$ &        0 &     16.07 &         0 &     16.99 \\ 
Struct$_{w}$   &     14.44 &     23.57 &     19.68 &     28.11 \\
Struct$_{l}$   &     14.81 & \bf 25.59 & \bf 20.97 & \bf 30.14 \\
Struct$_{e}$   & \bf 15.56 &     24.47 &     20.32 &     29.22
\end{tabular}
\caption{Results on queries failed by Elasticsearch.}
\label{table:unanswered-results}
\end{table}

\noindent To validate our hypothesis, we test structure matching on the subset of queries failed by Elasticsearch.
We first take the top-10 results from Elasticsearch then rerank the results using the scores from structure matching for queries that Elasticsearch gives P@1 of 0\%.
As shown in Table~\ref{table:unanswered-results}, structure matching is capable of reranking a significant portion (around 20\%) of these queries correctly, establishing that our hypothesis is true.

\subsection{Document Reranking}
\label{ssec:reranking}

The scores from Elastic$_{10}$ and Struct$_{*}$ for each document are fed into the binary classifier that decides whether or not to accept the top-1 result from Elasticsearch.
If not, the documents are reranked by the weighted sum of these scores (Section~\ref{subsection-document-reranking}).
The Rerank$_1$ row in Table~\ref{table:model-results} shows the results when all the weights = 1, which gives an over 4\% improvement of P@1 on the evaluation set.
The Rerank$_\lambda$ row shows the results when the optimized weights are used, which gives an additional 3\% boost on the development set but not on the evaluation set.

It is worth mentioning that we initially tackled this as a document classification task using convolutional neural networks similar to \newcite{kim:14a}; however, it gave P@1 $\approx$ 20\% and MRR $\approx$ 33\%.
Such poor results were due to the huge size of our documents, over 4.6K words on average, beyond the capacity of a CNN.
Thus, we decided to focus on reranking, which gave the best performance.

%% file: conclusion.tex
\section{Conclusion}
\label{conclusion-section}

We propose a cross-genre document retrieval task that matches between TV show transcripts and their descriptions in summaries and plots.
Our structure reranking approach gives an improvement of more than 4\% of P@1, showing promising results for this task.
In the future, we will add more structural information such as coreference relations to our structure matching and apply a more sophisticated parameter optimization technique such as the Bayesian optimization for finding $\lambda_*$.